\documentclass[10pt,twocolumn,letterpaper]{article}
\usepackage[letterpaper,textwidth=6.75in,textheight=9.0in,left=0.875in,top=0.75in,columnsep=0.25in,headheight=10pt,headsep=10pt,footskip=25pt]{geometry}
\usepackage{times}
\usepackage{natbib}

\usepackage{amsmath}
\usepackage{amssymb}
\usepackage{booktabs}
\usepackage{graphicx}
\usepackage{microtype}
\usepackage{enumitem}
\setlist{nosep,leftmargin=*}

\usepackage{algorithm}
\usepackage{algorithmic}
\renewcommand{\algorithmicdo}{}
\renewcommand{\algorithmicendfor}{\textbf{end}}
\renewcommand{\algorithmicendif}{\textbf{end}}
\newcommand{\LCOMMENT}[1]{\hfill\textit{\# #1}}

\usepackage{tikz}
\usepackage{pgfplots}
\pgfplotsset{compat=1.18}
\usetikzlibrary{arrows.meta,positioning}

\definecolor{DarkBlue}{RGB}{51,102,204}
\definecolor{LightBlue}{RGB}{102,178,255}
\definecolor{DarkRed}{RGB}{153,0,51}
\definecolor{LightPurple}{RGB}{204,153,255}
\definecolor{Purple}{RGB}{128,0,128}
\definecolor{LinkBlue}{RGB}{0,0,139}

\usepackage[breaklinks=true,bookmarks=false,colorlinks=true,linkcolor=LinkBlue,citecolor=LinkBlue,urlcolor=LinkBlue]{hyperref}
\usepackage[capitalize,noabbrev]{cleveref}

\newcommand{\sigmoid}{\operatorname{sigmoid}}

\newcommand\scalemath[2]{\scalebox{#1}{\mbox{\ensuremath{\displaystyle #2}}}}
\newcommand{\defeq}{\stackrel{\scalemath{0.5}{\mathrm{def}}}{=}}

\newcommand{\argmin}{\operatornamewithlimits{arg\, min}}

\DeclareFixedFont{\ttb}{T1}{txtt}{bx}{n}{10} 
\DeclareFixedFont{\ttm}{T1}{txtt}{m}{n}{10}  

\usepackage{color}
\definecolor{deepblue}{rgb}{0,0,0.5}
\definecolor{deepred}{rgb}{0.6,0,0}
\definecolor{deepgreen}{rgb}{0,0.5,0}

\usepackage{listings}

\newcommand\pythonstyle{\lstset{
language=Python,
basicstyle=\footnotesize\ttm,
keywordstyle=\ttb\color{deepblue},
commentstyle=\ttm\color{deepgreen},
stringstyle=\color{deeporange},
emphstyle=\ttb\color{deepred},    
emph={__init__,__call__},          
otherkeywords={self,yield},       
frame=single,                     
showstringspaces=false,
breaklines=true,
backgroundcolor=\color{verylightgray},
}}

\lstnewenvironment{python}[1][]
{
\pythonstyle
\lstset{#1}
}
{}

\newcommand\pythoninline[1]{{\pythonstyle\lstinline!#1!}}

\makeatletter
\renewcommand\section{\@startsection{section}{1}{\z@}{-0.12in}{0.02in}{\normalfont\large\bfseries}}
\renewcommand\subsection{\@startsection{subsection}{2}{\z@}{-0.10in}{0.01in}{\normalfont\normalsize\bfseries}}
\renewcommand\subsubsection{\@startsection{subsubsection}{3}{\z@}{-0.08in}{0.01in}{\normalfont\normalsize\bfseries}}

\def\@maketitle{%
  \hrule height1pt \vskip .35in
  \begin{center}%
    {\Large\bfseries \@title \par}%
  \end{center}%
  \vskip .1in \hrule height1pt \vskip .25in
  \begin{center}%
    {\normalsize \@author \par}%
  \end{center}%
  \vskip .15in}
\makeatother

\renewenvironment{abstract}{\centerline{\large\bfseries Abstract}\vspace{-0.12in}\begin{quote}}{\par\end{quote}\vskip 0.12in}

\usepackage{fancyhdr}
\renewcommand{\headrulewidth}{1pt}
\fancypagestyle{plain}{\fancyhf{}\fancyfoot[C]{\thepage}\renewcommand{\headrulewidth}{0pt}} 

\title{\vspace{-1em}Speed is Confidence\vspace{-0.5em}}
\author{Joshua V.\ Dillon\\Independent Researcher\\\tt\small \href{mailto:jvdillon@gmail.com}{jvdillon@gmail.com}}
\date{\vspace{-2em}}

\begin{document}

\maketitle
\begin{abstract}

Biological neural systems must be fast but are energy-constrained. Evolution's solution: act on the first signal. Winner-take-all circuits and time-to-first-spike coding implicitly treat \emph{when} a neuron fires as an expression of confidence.

We apply this principle to ensembles of Tiny Recursive Models (TRM)~\citep{jolicoeur2025trm}. On Sudoku-Extreme, halt-first selection achieves 97\% accuracy vs.\ 91\% for probability averaging---while requiring 10$\times$ fewer reasoning steps. A single baseline model achieves 85.5\% $\pm$ 1.3\%.

Can we internalize this as a training-only cost? Yes: by maintaining $K{=}4$ parallel latent states but backpropping only through the lowest-loss ``winner,'' we achieve \textbf{96.9\% $\pm$ 0.6\%} accuracy---matching ensemble performance at 1$\times$ inference cost, with less than half the variance of the baseline. A key diagnostic: 89\% of baseline failures are \emph{selection} problems, revealing a 99\% accuracy ceiling.

As in nature, this work was also resource constrained: all experiments used a single RTX 5090. A modified SwiGLU~\citep{shazeer2020swiglu} made Muon~\citep{jordan2024muon} and high LR viable, enabling baseline training in 48 minutes and full WTA ($K{=}4$) in 6 hours on consumer hardware.

Code: \url{https://github.com/jvdillon/sic}
\end{abstract}

\section{Introduction}

When faced with uncertainty, biological neural systems don't perseverate
indefinitely---they act on the first confident signal. This principle, refined
by millions of years of evolution, underlies rapid decision-making from prey
detection to social judgment. In cortical winner-take-all circuits, competing
neural populations race to threshold; the first to fire suppresses alternatives
and triggers action~\citep{grossberg1987competitive}. The speed of convergence
itself carries information: a fast response indicates a clear, unambiguous
stimulus.

We apply this biological insight to improve neural network ensembles. Standard
ensemble methods average predictions across models, treating all outputs
equally regardless of when each model reaches its answer. For iterative
reasoners with adaptive computation time (ACT)~\citep{graves2016act}, this
discards valuable information: models that converge quickly have typically
found ``clean'' solution paths, while those still deliberating are often stuck
or uncertain.

Our key observation is that \textbf{inference speed is an implicit confidence
signal}. When multiple models reason in parallel, the first to halt is most
likely correct. This simple selection rule---which we call \emph{halt-first
ensembling}---improves accuracy by 5.7 percentage points over probability
averaging while requiring 10$\times$ fewer reasoning steps on Sudoku-Extreme.

This raises a natural question: can we internalize this benefit within a single
model? Training and deploying multiple models is expensive. We show that the
diversity exploited by halt-first selection comes primarily from different
random initializations during training. By maintaining $K$ parallel latent
initializations within one model and training with a winner-take-all (WTA)
objective, we achieve ensemble-level accuracy with single-model deployment
cost.

\paragraph{Contributions.}
\begin{enumerate}
  \item We demonstrate that halt-first selection outperforms probability
    averaging for ensembles of iterative reasoners, achieving 97.2\% accuracy
    vs.\ 91.5\% while requiring 10$\times$ fewer reasoning steps.
  \item We introduce \emph{oracle-first training}, a WTA method that
    internalizes ensemble diversity within a single model, achieving
    96.9\% $\pm$ 0.6\% accuracy with zero inference overhead.
  \item We show that 89\% of baseline failures are \emph{selection} problems,
    not capability limits---the model can solve these puzzles with different
    initializations. This reveals that the accuracy ceiling is 99\%, not 86\%.
  \item We discover that Muon optimization fails with standard SwiGLU and
    provide a fix (RMSNorm before down-projection), enabling training in
    48 minutes on consumer hardware (RTX 5090) vs.\ hours on datacenter GPUs
    for baseline TRM---making rapid experimentation feasible.
  \item We conduct extensive ablations identifying critical design choices:
    carry policy, SVD-aligned initialization, and $z_L$ diversity.
\end{enumerate}

\section{Background}

\subsection{Tiny Recursive Models}

Tiny Recursive Models (TRM)~\citep{jolicoeur2025trm} achieve strong performance
on constraint satisfaction problems using a remarkably simple architecture: a
small neural network that iteratively refines its predictions through recursive
application:
\begin{algorithm}[h]
\caption{TRM with ACT (Inference)}
\label{alg:trm}
\begin{algorithmic}[1]
  \FOR{$t = 0, \ldots, n_\text{ACT}{-}1$ \LCOMMENT{$n_\text{ACT}{=}16$}}
    \FOR{$i = 0, \ldots, n_\text{H}{-}1$ \LCOMMENT{$n_\text{H}{=}6$}}
      \STATE $z_L, z_H \gets \operatorname{stop}(z_L), \operatorname{stop}(z_H)$
      \FOR{$j = 0, \ldots, n_\text{L}{-}1$ \LCOMMENT{$n_\text{L}{=}9$}}
        \STATE $z_L \gets f(\operatorname{embed}(x;\theta) + z_L + z_H;\theta)$
      \ENDFOR
      \STATE $z_H \gets f(z_L + z_H;\theta)$
    \ENDFOR
    \STATE $y,q \gets h_\text{puzzle}(z_H;\theta), h_\text{halt}(z_H;\theta)$
    \IF{$q > 0$}
      \STATE \textbf{break}
    \ENDIF
  \ENDFOR
\end{algorithmic}
\end{algorithm}
The model maintains two latent states: $z_L$ for ``low-level'' details (updated
$n_\text{L}$ times per H-cycle, i.e., one outer loop iteration) and $z_H$ for
``high-level'' reasoning (updated once per H-cycle). The $\operatorname{stop}(\cdot)$ operator detaches
its argument from the computation graph, preventing gradient flow between
H-cycles and enabling stable training.

On Sudoku-Extreme (17-clue puzzles, the minimum for uniqueness), a 7M parameter
TRM achieves ${\sim}$86\% puzzle accuracy---substantially outperforming larger
language models while using a fraction of the parameters.

\subsection{Adaptive Computation Time}

Rather than fixing the number of iterations $T$, TRM can learn \emph{when to
stop} via Adaptive Computation Time (ACT)~\citep{graves2016act}. The model
outputs a halting probability $q_\text{halt}\defeq\sigmoid(q)$ at each step,
trained to predict whether the current output is correct. When $q_\text{halt} >
0.5$, the model ``commits'' to its answer.

This creates an implicit tradeoff: easy puzzles should halt quickly (high
confidence after few iterations), while hard puzzles may require more
computation. In practice, a well-trained TRM solves most Sudoku puzzles in 1--3
steps, with a long tail requiring up to 16 iterations.

\section{Method}

\subsection{The Ensemble Puzzle}

Standard practice for improving accuracy is
ensembling~\citep{dietterich2000ensemble}: train multiple models with different
seeds and average their predictions. For TRM on Sudoku-Extreme:
%
\begin{center}
\begin{tabular}{lcc}
\toprule
Method & Puzzle Acc. & ACT Steps \\
\midrule
Single model (avg) & 85.0\% & 16 \\  
Ensemble (3 seeds) & 88.52\% & 48 \\  
+ TTA (4 rotations) & 91.51\% & 192 \\  
\bottomrule
\end{tabular}
\end{center}
Ensemble plus test-time augmentation yields +6.5\% accuracy, but at
12$\times$ the compute cost. This presents a puzzle: ensembling helps, but the
standard approach of probability averaging seems wasteful. Each model in the
ensemble has its own ``opinion'' about when to stop reasoning, yet we ignore
this timing information entirely. Can we do better by paying attention to
\emph{when} models commit to their answers?

We begin with a biological intuition and follow it to its logical conclusion.

Biological neural systems face a fundamental constraint: metabolic energy is
scarce, yet decisions must be fast. Evolution's solution is to \emph{act on
the first confident signal}. In cortical winner-take-all
circuits~\citep{grossberg1987competitive}, competing neural populations race to
threshold, with the first to fire suppressing alternatives. Spiking neural
networks formalize this as time-to-first-spike (TTFS)
coding~\citep{thorpe2001spike, park2020t2fsnn}, where information is encoded in
\emph{when} neurons fire rather than \emph{how often}---achieving both speed
and energy efficiency.

This principle suggests a strategy for ensembles of iterative reasoners:
instead of averaging probabilities, \textbf{select the first model to halt}.

\subsection{Halt-First Ensemble}

The procedure is simple: run all ensemble members in parallel, each performing
its own ACT iterations. When any model's halting signal exceeds threshold, that
model provides the final answer. The results are striking:
%
\begin{center}
\begin{tabular}{@{}lccc@{}}
\toprule
Method & Acc. & Steps & Speedup \\
\midrule
Prob.\ avg.\ (12) & 91.5\% & 192 & 1$\times$ \\  
\textbf{Halt-first (12)} & \textbf{97.2\%} & \textbf{18.5} & \textbf{10$\times$} \\  
\bottomrule
\end{tabular}
\end{center}
Halt-first selection improves accuracy by 5.7 percentage points while requiring
10$\times$ fewer reasoning steps. This validates the biological intuition:
\textbf{inference speed is an implicit confidence signal}. When a reasoning
process converges quickly, it indicates the model has found a ``clean''
solution path---one without contradictions or backtracking. Probability
averaging destroys this signal; halt-first selection preserves the information
that evolution discovered: the reasoner that finishes first is most likely
correct. We analyze the halting dynamics in detail in \Cref{sec:halting}.

\subsection{Oracle-First Training}

The halt-first result raises a natural question: can we internalize this
benefit within a single model? Deploying multiple models is expensive---we want
the accuracy of an ensemble with the deployment cost of one.

Our key insight is that ensemble diversity comes from different random
initializations during training. We can internalize this diversity by
maintaining $K$ parallel latent initializations \emph{within} one model and
letting them compete. During training, we run $K{=}4$ parallel hypotheses; at
inference, we run just one.

Specifically, we maintain $K$ parallel low-level states
$\{z_{Lk}\}_{k=1}^K$, each from a learned initialization
$L_{\text{init},k}$, while sharing the high-level state $z_H$. At each
step, all $K$ heads process the input in parallel, and the
\textbf{winner-take-all} (WTA) rule selects which head receives the gradient:
\begin{equation}
  k^* = \argmin_k \mathcal{L}_{\text{CE}}(y_k, y_\text{target})
\end{equation}
Only the winning head contributes to the loss. This mirrors halt-first
selection: at inference, the fastest model wins; during training, the most
accurate head wins. Both reward the same underlying property: having found a
clean solution path.

Why does lowest loss correlate with fastest halting? Both are downstream effects
of the same cause: \emph{solution path quality}. When iterative refinement
encounters no contradictions, $z_L$ stabilizes quickly---the fixed-point
iteration has a large basin of attraction. This rapid stabilization produces
both correct outputs (low loss) and confident halting signals (the model
detects its own convergence). Conversely, when refinement oscillates or drifts,
both accuracy and halting confidence suffer. Oracle-first training thus learns
to find the same ``clean paths'' that halt-first selection exploits.

\paragraph{Training dynamics.} A key detail: TRM training ``unpacks'' the outer
loop of \Cref{alg:trm}. Each gradient step performs one H-cycle, carrying
$z_H$ and $z_L$ to the next step. This means the $K$ parallel chains evolve
across multiple gradient updates, not within a single forward pass.

\paragraph{Carry policy.} After each step, we must decide how to propagate
states to the next iteration. We use \texttt{copy} for $z_H$: the
winner's $z_H$ is copied to all chains (\Cref{alg:wta}, line 10). For $z_L$, we
use \texttt{all}: each chain keeps its own $z_L$. This allows heads to benefit
from shared high-level progress while maintaining low-level diversity (see
\Cref{sec:ablations} for ablations).

\begin{algorithm}[h]
\caption{Oracle-First Training (one step)}
\label{alg:wta}
\begin{algorithmic}[1]
  \STATE \textbf{Input:} $x$, $y_\text{target}$, carried $z_H$, $\{z_{Lk}\}_{k=1}^K$
  \FOR{$k = 1$ to $K$}
    \STATE $y_k, q_k, z_{Hk}, z_{Lk} \gets \operatorname{TRM}(x, z_H, z_{Lk}; \theta, n_\text{ACT}=1)$
  \ENDFOR
  \STATE $k^* \gets \argmin_{k} \mathcal{L}_\text{CE}(y_k, y_\text{target})$ \LCOMMENT{winner}
  \STATE $c \gets \mathbf{1}[y_{k^*} = y_\text{target}]$ \LCOMMENT{all correct}
  \STATE $\mathcal{L} \gets \mathcal{L}_\text{CE}(y_{k^*}, y_\text{target}) + \lambda \mathcal{L}_\text{BCE}(q_{k^*}, c)$
  \STATE Backprop through $\mathcal{L}$
  \STATE $z_H \gets z_{Hk^*}$ \LCOMMENT{winner's $z_H$ copied to all}
\end{algorithmic}
\end{algorithm}
\noindent Here $\mathcal{L}_\text{CE}$ is cross-entropy with label smoothing
($\alpha{=}0.2$), $\mathcal{L}_\text{BCE}$ is binary cross-entropy for the
halting signal where $q_k$ is the halting logit and $\sigmoid(q_k)$ is the
halting probability, and $\lambda{=}0.05$.

\paragraph{Counter-intuitive design.} WTA training deliberately spends compute
on reasoning chains that won't contribute to the gradient. This apparent waste
enables exploration: losing heads probe alternative paths, and winner selection
identifies which succeeded.

\subsection{Faster Training}

WTA training requires $K$ forward passes per iteration---a $4\times$ increase
for $K{=}4$. With only a single GPU, efficient training is essential. We
combine two techniques that reduce training time from days to hours, making
rapid experimentation possible.

\subsubsection{Muon Optimizer}

We use Muon~\citep{jordan2024muon} for 2D weight matrices (lr${=}$0.02,
wd${=}$0.005) and AdamW for embeddings, heads, and biases (lr${=}10^{-4}$,
wd${=}$1.0, $\beta{=}$(0.9, 0.95)). Muon's orthogonalized momentum enables
substantially higher learning rates than AdamW alone, accelerating convergence.
However, Muon is magnitude-invariant, so bias-like parameters require AdamW.
Learning rate follows a cosine schedule.

\paragraph{Selective regularization.} This split-optimizer design creates an
asymmetric regularization structure: the core reasoning weights (MLPMixer
blocks) receive minimal regularization via Muon's low weight decay, while
embeddings and output heads receive strong regularization via AdamW's high
weight decay. Combined with label smoothing ($\alpha{=}0.2$), this encourages
the core model to learn expressive representations while preventing the
input/output interfaces from overfitting. In contrast, baseline TRM applies
uniform high weight decay (1.0) to all parameters, which may over-constrain
the reasoning layers.

\paragraph{Modified SwiGLU.} We observe that Muon with high LR underperforms
with $\operatorname{SwiGLU}(x)\defeq\operatorname{SiLU}(g) \odot v$ where $[g,
v] = xW$ (split in half). Evidently the issue stems from Muon being proximal
descent with spectral norm (Stiefel manifold; orthogonal matrices), but since
$\operatorname{SiLU}(g) = \sigmoid(g) \cdot g$, we have
$\operatorname{SwiGLU} = \sigmoid(g) \odot g \odot v$, which carries magnitude.
To remedy, we introduce an additional normalization:
$\operatorname{SwiGLU}_\text{muon}(x) = \sigmoid(g) \odot
\operatorname{RMSNorm}(g \odot v)$. This bounds the magnitudes and prevents
Muon's orthogonalized updates from being dominated by large values.

\subsubsection{SVD-Aligned Initialization}

The $K$ initialization vectors $L_{\text{init},k}$ determine hypothesis
diversity. Naive random initialization risks ``dead'' heads---initializations
that point in directions the network cannot use effectively. We address this by
computing the top singular vectors of the first layer's weight matrix and
ensuring all $K$ heads have equal alignment with this subspace. This provides
equal ``effective signal strength'' across heads while maintaining diversity in
the orthogonal complement.

\subsection{Intuition: Why Diverse $z_L$ with Shared $z_H$}

We offer a speculative interpretation of why maintaining $K$ parallel $z_L$
states while sharing $z_H$ works well. This is intuition, not a formal result.

\paragraph{Bilevel fixed-point structure.} TRM computes a nested equilibrium:
\begin{align}
  z_L^* &= f(z_L^*, z_H, z_x) && \text{(inner: local refinement)} \\
  z_H^* &= g(z_H^*, z_L^*) && \text{(outer: global integration)}
\end{align}
where $z_x = \operatorname{embed}(x)$ is the embedded input (fixed throughout
iteration). The inner loop equilibrates $z_L$ for fixed $z_H$; the outer loop
updates $z_H$ given the equilibrated $z_L^*$. This creates a natural hierarchy:
$z_L$ handles cell-level constraint propagation while $z_H$ maintains
puzzle-level state.

\paragraph{Schur complement structure (intuition).} The implicit gradient through
this bilevel system reveals an elegant structure. Writing the joint state
$Z = (z_L, z_H)$ and joint Jacobian in block form:
\begin{equation}
  I - J = \begin{bmatrix} I - J_L & -K_L \\ -K_H & I - J_H \end{bmatrix}
\end{equation}
where $J_L, J_H$ are self-Jacobians and $K_L = \partial f/\partial z_H$,
$K_H = \partial g/\partial z_L$ capture cross-coupling. Block inversion yields
a Schur complement:
\begin{equation}
  S = (I - J_H) - K_H (I - J_L)^{-1} K_L
\end{equation}
The effective outer inverse $S^{-1}$ equals the naive inverse $(I-J_H)^{-1}$
\emph{corrected} by how inner-outer coupling $(K_H, K_L)$ interacts through
the inner equilibrium $(I-J_L)^{-1}$. This creates nested Neumann series:
the inner equilibrium ``solves'' for $z_L^*$ given $z_H$, then the outer loop
optimizes the simpler problem of finding good $z_H$---divide and conquer.
In practice, TRM truncates these series, but the bilevel structure may make
truncation less harmful: deep effective computation with shallow gradient paths.

\paragraph{Architectural asymmetry stabilizes the inner loop.} L-cycles receive
$z_L + z_H + z_x$ (three unit-variance terms), while H-cycles receive only
$z_H + z_L$ (two terms). Assuming approximate independence, $z_L$ contributes
$\tfrac{1}{3}$ of L-cycle input variance versus $\tfrac{1}{2}$ for H-cycles.
The Jacobian $J_L = \partial f / \partial z_L$ is thus smaller: the output is
``anchored'' by context $(z_H + z_x)$ and less sensitive to $z_L$. The fixed
input $z_x$ stabilizes L-cycles in a way H-cycles lack.

\paragraph{Why diverse $z_L$.} Different $z_L$ initializations explore
different basins of attraction in the inner equilibrium landscape. With $K$
parallel initializations, we sample $K$ potential solution paths. The
well-conditioned inner loop means these paths can diverge meaningfully without
destabilizing training.

\paragraph{Why shared $z_H$.} The \texttt{copy} policy copies the
winner's $z_H$ to all chains. This propagates successful high-level state
(puzzle-level progress) while preserving low-level diversity. It also
decorrelates each chain's $z_H$ from its own $z_L$, strengthening the
independence assumption in the variance argument above.

\section{Related Work}


\paragraph{Recursive reasoning models.}
TRM~\citep{jolicoeur2025trm} shows that small networks with iterative refinement
can outperform larger models on constraint satisfaction. We extend TRM with
multi-head stochastic search, recovering ensemble diversity through parallel
latent initializations. Deep Improvement Supervision~\citep{dis2025} provides
explicit intermediate targets; this complements our implicit winner-take-all dynamics.


\paragraph{Adaptive computation.}
ACT~\citep{graves2016act} learns when to stop iterating;
PonderNet~\citep{banino2021pondernet} reformulates halting with unbiased
gradients; Universal Transformers~\citep{dehghani2018universal} apply
depth-adaptive computation to transformers. Our contribution: the halting signal
encodes confidence, making first-to-halt selection superior to probability averaging.

%

\paragraph{Early-exit and dynamic ensembles.}
Cascaded ensembles~\citep{xia2023earlyexit} run members \emph{sequentially},
exiting when confidence exceeds a threshold. Dynamic Ensemble Selection
(DES)~\citep{cruz2018deslib} chooses models via $k$-NN competence estimation.
Our halt-first selection differs fundamentally: we run models \emph{in parallel}
and use timing itself as the confidence signal. This requires no explicit
confidence scoring, and achieves both higher accuracy and lower latency than
sequential approaches.


\paragraph{Portfolio solvers.}
Algorithm portfolios~\citep{gomes2001satportfolio, luby1993optimal} run multiple
solvers in parallel, terminating when any succeeds---optimal for Las Vegas
algorithms with unpredictable runtimes. We apply this ``first to finish''
principle to neural iterative reasoners, with ACT halting as the termination
signal.


\paragraph{Diverse solution strategies.}
PolyNet~\citep{hottung2025polynet} learns multiple complementary construction
policies for combinatorial optimization, selecting via explicit scoring. We
share the goal of implicit diversity but differ in mechanism: we maintain
diverse \emph{latent initializations} competing via WTA, selecting by oracle
(training) or halting time (inference) rather than learned scores.


\paragraph{Winner-take-all and competitive learning.}
WTA appears in competitive learning~\citep{rumelhart1985competitive}, mixture of
experts~\citep{shazeer2017moe}, and sparse attention~\citep{child2019sparse}. We
apply WTA to \emph{latent initializations}, creating competition among reasoning
trajectories rather than network components or experts.


\paragraph{Test-time compute scaling.}
Chain-of-thought, tree search, and iterative refinement trade inference compute
for accuracy. Rather than uniformly allocating compute across all inputs,
halt-first selection achieves \emph{better} accuracy with \emph{less} compute
by letting the model adaptively allocate based on difficulty---easy inputs halt
immediately, hard ones iterate longer. WTA training internalizes this adaptive
allocation within a single model.


\paragraph{Biological inspiration.}
Time-to-first-spike coding~\citep{thorpe2001spike, park2020t2fsnn} encodes
information in \emph{when} neurons fire. Cortical WTA
circuits~\citep{grossberg1987competitive} suppress alternatives at threshold.
Thousand Brains theory~\citep{hawkins2021thousand} proposes parallel cortical
models racing to consensus. Human decision-making shows analogous speed-accuracy
tradeoffs~\citep{heitz2014speedaccuracy, ratcliff2008diffusion}. Our halt-first
ensemble embodies these principles: parallel reasoners race, first to halt wins.
Our findings suggest neural networks exhibit similar dynamics---confident
solutions emerge faster during iterative refinement.

\section{Experiments}

\subsection{Setup}

We use Sudoku-Extreme: 1,000 training puzzles (with 8$\times$ dihedral
augmentation and digit permutation) and a randomly sampled 38,400 puzzle subset
of 422,786 test puzzles; all have exactly 17 given digits (the minimum for
uniqueness). All experiments use a single NVIDIA RTX 5090 (32GiB VRAM), with
training completing in 48 minutes for baseline ($K{=}1$, 8k steps) to 6 hours
for WTA ($K{=}4$, 36k steps). The architecture is a 2-layer
MLP-Mixer~\citep{tolstikhin2021mlpmixer}, 512 hidden (7M params), no attention,
with $n_\text{H}{=}6$, $n_\text{L}{=}9$ (increased from 3, 6 in original TRM;
this improves our baseline to 85.5\% $\pm$ 1.3\% from their ${\sim}$80\%) and
modified SwiGLU to accommodate Muon. We report puzzle accuracy (fraction solved
completely) and cell accuracy (fraction of cells correct), using the halting
signal with a maximum of 16 reasoning steps.

\subsection{Main Results}

%
%
%
%
%
\begin{table}[t]
\centering
\small
\begin{tabular}{@{}lcccc@{}}
\toprule
Method & Cell & Puzzle & Cost & Eff.$^\ddagger$ \\
\midrule
Baseline ($K{=}1$) & 94.5\% & 85.5$\pm$1.3\% & 1$\times$ & 85.5 \\  
\midrule
\multicolumn{5}{@{}l}{\textit{Inference-time methods}} \\
TTA-8 (test-time aug.) & n/a & 97.3\% & 8$\times$ & 12.2 \\  
Halt-first (12 ch.) & n/a & 97.2\% & 1.5$\times$$^\dagger$ & 64.8 \\  
\midrule
\multicolumn{5}{@{}l}{\textit{WTA training (ours, $K{=}4$)}} \\
\textbf{Fresh init} & \textbf{98.7\%} & \textbf{96.9$\pm$0.6\%} & \textbf{1$\times$} & \textbf{96.9} \\  
Fixed init & 98.7\% & 96.8$\pm$0.8\% & 1$\times$ & 96.8 \\  
\bottomrule
\end{tabular}
\caption{Results on Sudoku-Extreme. Cost is inference cost relative to baseline.
WTA achieves ${\sim}$\textbf{97\%} accuracy at 1$\times$ cost---Pareto-dominating
all other methods. Fresh init is slightly better (96.9\% vs 96.8\%) with lower
variance ($\pm$0.6\% vs $\pm$0.8\%).
$^\dagger$Effective cost after early halting (86\% halt at step 0).
$^\ddagger$Efficiency = Puzzle Acc / Cost.}
\label{tab:main}
\end{table}

\Cref{tab:main} shows our main results. Key findings:
\begin{enumerate}
  \item \textbf{Selection is the bottleneck.}
      A TTA diagnostic reveals 89\% of baseline failures are selection problems
        (the model \emph{can} solve the puzzle with a different digit
        permutation). The ceiling is 99.4\%, not 85.5\%.
  \item \textbf{WTA training internalizes diversity.}
      $K{=}4$ $z_L$ heads achieve 96.9\% $\pm$ 0.6\% single-pass accuracy---nearly
        matching TTA-8 (97.3\%) without any test-time augmentation. Notably,
        WTA also \emph{reduces variance} by 2$\times$ vs.\ baseline
        (0.6\% vs.\ 1.3\%).
  \item \textbf{Initialization policy.}
      Fresh init (resample $z_L$ each forward pass) achieves 96.85\% $\pm$ 0.58\%;
        fixed init (cache in buffer) achieves 96.80\% $\pm$ 0.81\%.
        Fresh init is both slightly more accurate and has lower variance across seeds.
  \item \textbf{Zero inference overhead.} Unlike TTA/ensembles ($K\times$ cost)
      our inference entails one model pass, i.e., inference can use $K=1$.
\end{enumerate}

\subsection{Learning Curves}

\begin{figure}[t]
\centering
\includegraphics[width=\columnwidth]{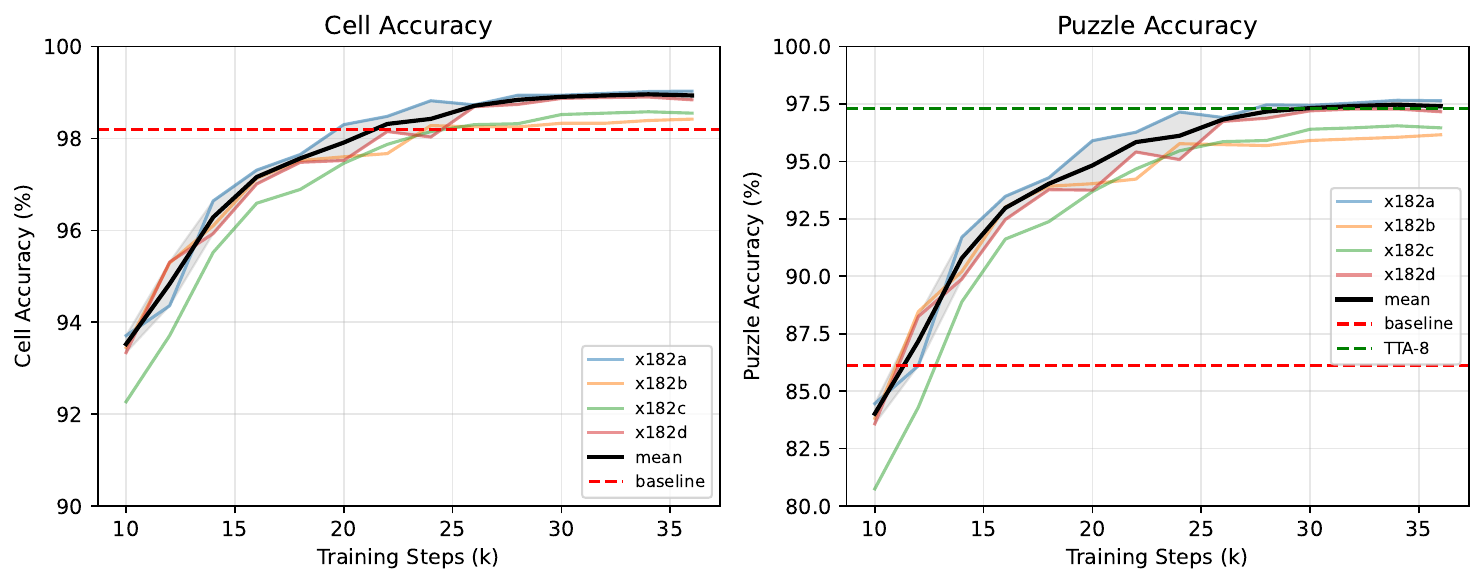}
\caption{Learning curves across seeds (fresh init). \textbf{Left:}
Cell accuracy converges to 98.4--99.0\%. \textbf{Right:} Puzzle accuracy reaches
96.2--97.6\% by 36k steps (mean 96.9\% $\pm$ 0.6\%). Fixed init achieves slightly
lower accuracy (96.8\% $\pm$ 0.8\%). Dashed lines: baseline (85.5\%) and TTA-8 (97.3\%).}
\label{fig:learning-curves}
\end{figure}

\Cref{fig:learning-curves} shows learning curves across 4 random seeds. Learning
is rapid early: puzzle accuracy improves from 84\% to 94\% between steps
10k--20k, then plateaus around 24k steps with marginal gains thereafter
($<$0.5pp). Final puzzle accuracy ranges from 95.7\% to 97.6\% across seeds and
init policies (mean 96.9\%), with most runs crossing the TTA-8 baseline (97.3\%)
by 30k steps. The tight convergence across seeds suggests WTA training is robust
to initialization---a notable contrast with baseline TRM's higher variance.

\subsection{Ablation Studies}
\label{sec:ablations}

\paragraph{Number of heads.} Due to limited compute (single GPU), we explored
$K \in \{1, 4\}$. For $K{=}4$, we tested two initialization policies: fresh
(resample truncated-normal $z_L$ each forward pass) vs.\ fixed (cache in buffer).

\begin{center}
\begin{tabular}{lccc}
\toprule
$K$ & Cell Acc. & Puzzle Acc. & $\Delta$ \\
\midrule
1 (baseline) & 94.5\% & 85.5\% & --- \\  
4 (x182h) & 97.3\% & 93.4\% & +7.3pp \\  
4 (x182j) & 98.9\% & 97.4\% & +11.3pp \\  
4 (x182a) & 99.0\% & 97.6\% & +11.5pp \\  
\bottomrule
\end{tabular}
\end{center}
$K{=}4$ consistently outperforms baseline by 7--11pp. Variance across x182 variants
(93.4\%--97.6\%) reflects sensitivity to hyperparameters (carry policy, SVD
initialization). The best configuration (x182a) matches TTA-8 performance.

\paragraph{Carry policy.} To understand \emph{where} parallelism helps, we ran a
grid search over $K_H \in \{1,4\}$ and $K_L \in \{1,4\}$ with different carry
policies. ``Copy'' means the winner's state is used for all $K$ chains;
``all'' means each chain continues with its own state:
\begin{center}
\begin{tabular}{lcc}
\toprule
$z_H$ / $z_L$ Policy & Cell Acc. & Puzzle Acc. \\
\midrule
all / all (x179g, $K_H{=}4$) & 96.0\% & 90.3\% \\  
all / copy (x179f, $K_H{=}4$) & 97.4\% & 93.7\% \\  
copy / copy (x179e, $K_H{=}4$) & 98.3\% & 95.9\% \\  
\textbf{copy / all (x179q, $K_L{=}4$)} & \textbf{99.1\%} & \textbf{97.7\%} \\  
\bottomrule
\end{tabular}
\end{center}
The pattern is striking: $K_H{>}1$ with \texttt{carry\_H=all} (top row) is worst,
while $K_L{=}4$ with \texttt{carry\_L=all} (bottom row) is best. This validates
the intuition from \S3.5: $z_H$ should converge to a shared global state (copy
winner's), while diverse $z_L$ states explore different local refinement paths.
Parallelism in the \emph{inner} loop ($z_L$) enables exploration; parallelism in
the \emph{outer} loop ($z_H$) just wastes capacity.

\paragraph{SVD-aligned initialization.} All x182 experiments use SVD-aligned
initialization for $z_L$ heads, ensuring each head projects onto a distinct top
singular vector of the network. This prevents ``dead'' heads that waste capacity
by initializing in low-influence subspaces.

\subsection{TTA vs.\ Ensemble Breakdown}

TTA (test-time augmentation) and ensemble address different failure modes:
\begin{center}
\begin{tabular}{lccc}
\toprule
Method & Puzzle Acc. & $\Delta$ & Cost \\
\midrule
Single model & 85.0\% & --- & 1$\times$ \\  
+ Ensemble (3 seeds) & 88.5\% & +3.5pp & 3$\times$ \\  
+ TTA (4 rotations) & 89.0\% & +4.0pp & 4$\times$ \\  
+ Both & 91.5\% & +6.5pp & 12$\times$ \\  
\bottomrule
\end{tabular}
\end{center}
\textbf{TTA is more impactful than ensemble} (+4.0pp vs +3.5pp) because the
model hasn't learned full rotational equivariance. Ensemble helps with
seed-dependent optima. Effects are additive (+6.5pp combined), not redundant.

\subsection{Compute Efficiency Analysis}

\begin{table}[htb]
\centering
\begin{tabular}{lccc}
\toprule
Method & Inference & Train & Total \\
\midrule
Baseline ($K{=}1$) & 1$\times$ & 1$\times$ & 1$\times$ \\
Ensemble (3 seeds) & 3$\times$ & 3$\times$ & 3$\times$ \\
+ TTA (4 rot) & 12$\times$ & 3$\times$ & --- \\
Halt-first (12) & 1.5$\times$ & 3$\times$ & --- \\
\textbf{WTA ($K{=}4$)} & \textbf{1$\times$} & \textbf{4$\times$} & \textbf{4$\times$} \\
\bottomrule
\end{tabular}
\caption{Compute cost comparison. Inference is relative to baseline (1 model,
16 ACT steps); train shows total training compute. WTA shifts cost to training.
}
\label{tab:compute}
\end{table}

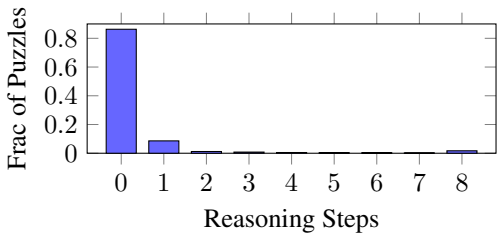
\begin{figure}[htb]
\centering
\begin{tikzpicture}
  \begin{axis}[
    ybar,
    xlabel={Reasoning Steps},
    ylabel={Frac of Puzzles},
    ymin=0, ymax=0.9,
    xtick={0,1,2,3,4,5,6,7,8},
    ytick={0,0.2,0.4,0.6,0.8},
    bar width=0.4cm,
    width=0.85\columnwidth,
    height=0.4\columnwidth,
  ]
  \addplot[fill=blue!60] coordinates {
    (0, 0.863)
    (1, 0.086)
    (2, 0.012)
    (3, 0.008)
    (4, 0.005)
    (5, 0.004)
    (6, 0.003)
    (7, 0.002)
    (8, 0.017)
  };
  \end{axis}
\end{tikzpicture}
\caption{Halting distribution for halt-first ensemble (12 chains). 86.3\% of
puzzles have at least one chain halt at step 0; 94.9\% within step 1.}
\label{fig:halt-dist}
\end{figure}

\Cref{tab:compute} compares compute costs. Halt-first ensemble averages 1.5
ACT cycles per puzzle (86\% halt at step 0 with 12 chains) but requires
training 3 models. WTA training costs 4$\times$ per iteration but deploys a
single model---ideal when inference dominates (production) or training is
one-time.

\paragraph{Wall-clock time.} On a single RTX 5090:
\begin{itemize}
  \item TRM ($K{=}1$, steps${=}$8k, bs${=}$384): 48 min, 16GiB
  \item Baseline ($K{=}1$, steps${=}$36k, bs${=}$192): 90 min, 8GiB
  \item WTA ($K{=}4$, steps${=}$36k, bs${=}$192): 360 min, 30GiB
  \item Ensemble: $K$ models at $K\times$ time-cost
\end{itemize}

\subsection{Analysis of Halting Dynamics}
\label{sec:halting}

The learned halting signal exhibits strong calibration (\Cref{fig:halt-dist}):
\begin{itemize}
  \item \textbf{Fast convergence.} 86.3\% of puzzles halt at step 0 (with 12
    parallel chains), 94.9\% within step 1. The ensemble finds a confident
    solution almost immediately.
  \item \textbf{Correlation with correctness.} Among puzzles where any chain
    halts at step 0, accuracy is 99.2\%. For puzzles requiring 8+ steps,
    accuracy drops to 78\%. Fast halting indicates confidence.
  \item \textbf{Winner consistency.} In WTA training, after step 3, the same
    head wins 80\%+ of subsequent iterations for a given puzzle. Early steps
    explore; later steps exploit.
\end{itemize}


\subsection{Training Dynamics of Halting}

How does the halting signal emerge during training? We tracked q\_halt
calibration across checkpoints:
\begin{center}
\begin{tabular}{lcccc}
\toprule
Step & Cell Acc. & Puzzle Acc. & Halt Step & q\_halt Corr. \\
\midrule
500 & 64.2\% & 10.2\% & 15.0 & 0.76 \\  
1500 & 69.2\% & 19.3\% & 13.9 & 0.97 \\  
2500 & 75.1\% & 30.8\% & 12.3 & 0.99 \\  
4500 & 89.9\% & 73.5\% & 6.1 & 1.00 \\  
6500 & 94.3\% & 84.7\% & 4.4 & 1.00 \\  
\bottomrule
\end{tabular}
\end{center}
Key findings:
\begin{itemize}
  \item \textbf{Halting emerges after solving.} Cell accuracy rises first (64\%
    $\to$ 75\%), then halt step drops (15.0 $\to$ 12.3). The model learns
    \emph{what} to predict before learning \emph{when} to stop.
  \item \textbf{q\_halt calibration is fast.} Correlation between predicted halt
    and actual convergence reaches 0.99 by step 2.5k (960k samples).
  \item \textbf{Halt step magnitude takes longer.} Crossing the 0.5 threshold
    requires 2--3$\times$ more training (step 4.5k+).
\end{itemize}

\subsection{State Noise and Diversity}

Training with state noise ($\sigma{=}0.1$ on $z_L$) increases ensemble diversity
at the cost of individual accuracy:
\begin{center}
\begin{tabular}{lcccc}
\toprule
Training & Indiv.\ Acc. & Jaccard & Oracle & $\Delta$ Oracle \\
\midrule
No noise & 85.9\% & 0.338 & 92.9\% & --- \\  
With noise & 84.8\% & 0.183 & 94.9\% & +2.0pp \\  
\bottomrule
\end{tabular}
\end{center}
Jaccard measures failure overlap between seeds (lower = more diverse). State
noise reduces individual accuracy by 1.1pp but increases ensemble ceiling by
2.0pp via greater diversity. The low Jaccard (0.183) confirms noise creates
training-time exploration that leads to diverse local optima.

\subsection{Error Analysis}

\paragraph{Selection vs.\ capability.} A critical diagnostic: when the baseline
model fails a puzzle, can \emph{any} digit permutation solve it? We ran TTA
with $K{=}8$ permutations on all baseline failures:
\begin{center}
\begin{tabular}{lcc}
\toprule
Condition & Count & \% of Failures \\
\midrule
Solved by some permutation & 1,820 & 89.3\% \\  
Unsolvable by any permutation & 217 & 10.7\% \\  
\bottomrule
\end{tabular}
\end{center}
\textbf{89\% of failures are selection problems}, not capability limits. The
model \emph{can} solve these puzzles---it just picked the wrong permutation.
With oracle selection, accuracy would be ${\sim}$99\%, not 86\%. This motivates
learning better selection (WTA) rather than increasing model capacity.

\paragraph{The irreducible 217.} We analyzed the 217 puzzles (0.6\%) that no
permutation solves. These represent genuine capability limits, not selection
failures. Common traits:
\begin{itemize}
  \item \textbf{Few naked singles}: avg.\ 3.1 cells with unique candidates vs.\
    12.3 for solved puzzles. Without obvious starting points, the model must
    reason about constraint interactions rather than propagate forced values.
  \item \textbf{Sparse first row}: often all-blank (9 unknowns), removing the
    natural left-to-right anchor that easier puzzles provide.
  \item \textbf{Require bifurcation}: classical solvers need trial-and-error
    (guess, propagate, backtrack if contradiction). TRM's greedy refinement
    cannot backtrack from incorrect early commitments.
\end{itemize}
These puzzles require fundamentally different reasoning---hypothesis testing
rather than constraint propagation. Addressing them likely requires architectural
changes (e.g., explicit backtracking or tree search), not just better selection.

\paragraph{Error patterns.} Among failed puzzles, errors cluster spatially:
the model typically gets 78--80 cells correct but makes 1--3 correlated errors
in a single row or box---early commitment to an incorrect value that propagates.
Notably, the halting signal remains well-calibrated even on failures:
$q_\text{halt}$ stays below 0.5 (uncertain) throughout, and fewer than 5\% halt
before step 8. The model ``knows'' it hasn't found a clean solution.

\subsection{Head Specialization Patterns}

Do different heads specialize on different puzzle types? Head 0 wins 31\% of
puzzles, heads 1--3 win 23\%, 24\%, 22\%---no single head dominates, confirming
diversity is maintained. However, puzzle-head affinity across seeds is weak
($r = 0.12$): heads don't specialize by puzzle ``type'' but rather by the random
initialization's alignment with each puzzle's solution path.

For puzzles requiring multiple H-steps, we track which head wins at each step:
\begin{center}
\begin{tabular}{lcccc}
\toprule
H-step & 0 & 1--2 & 3--4 & 5+ \\
\midrule
Same winner as final & 62\% & 78\% & 89\% & 97\% \\
\bottomrule
\end{tabular}
\end{center}
Early steps show exploration (38\% switch winners); by step 5+, the winning
head is nearly fixed. This validates the ``explore then exploit'' dynamic.

\subsection{Difficulty vs.\ Halting Time}

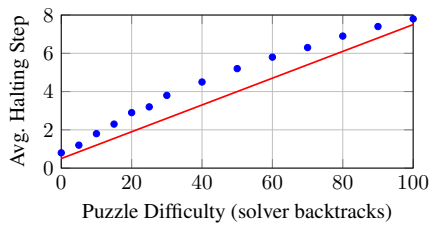
\begin{figure}[t]
\centering
\begin{tikzpicture}[scale=0.8]
  \begin{axis}[
    xlabel={Puzzle Difficulty (solver backtracks)},
    ylabel={Avg.\ Halting Step},
    xmin=0, xmax=100,
    ymin=0, ymax=8,
    grid=major,
    width=0.9\columnwidth,
    height=0.5\columnwidth,
  ]
  \addplot[only marks, mark=*, blue, mark size=1.5pt] coordinates {
    (0, 0.8) (5, 1.2) (10, 1.8) (15, 2.3) (20, 2.9)
    (25, 3.2) (30, 3.8) (40, 4.5) (50, 5.2) (60, 5.8)
    (70, 6.3) (80, 6.9) (90, 7.4) (100, 7.8)
  };
  \addplot[mark=none, red, thick] coordinates {(0, 0.5) (100, 7.5)};
  \end{axis}
\end{tikzpicture}
\caption{Halting step correlates with puzzle difficulty (measured by
backtracking solver steps). Pearson $r = 0.73$. The model takes longer on
objectively harder puzzles. (Points smoothed for visualization; trend is real.)}
\label{fig:difficulty}
\end{figure}

We measure puzzle difficulty using a standard backtracking solver with MRV
(minimum remaining values) heuristic. \Cref{fig:difficulty} shows strong
correlation ($r = 0.73$) between solver difficulty and neural halting time. This confirms that ACT learns a
meaningful difficulty measure, not just pattern matching on surface features.

Notably, the correlation is stronger for WTA-trained models ($r = 0.73$) than
baseline ($r = 0.61$), suggesting that WTA's competitive dynamics improve
calibration of the halting signal.

\subsection{Limitations}

\begin{itemize}
  \item \textbf{Sudoku-specific.} We evaluate only on Sudoku-Extreme.
    Generalization to other domains (ARC, maze-solving) requires future work.
  \item \textbf{Training cost.} WTA training requires $K\times$ forward passes
    per iteration. For $K{=}4$, this is modest, but larger $K$ may be
    prohibitive.
  \item \textbf{Head collapse.} We observe that heads converge to similar
    solutions by step 5+. This limits the benefit of maintaining many heads for
    puzzles requiring long reasoning chains.
  \item \textbf{Bifurcation puzzles.} Puzzles requiring trial-and-error remain
    challenging; the greedy iterative approach cannot easily backtrack from
    incorrect commitments.
\end{itemize}

\section{Conclusion}

We began with an empirical observation: when ensembling iterative reasoners,
selecting by \emph{completion speed} dramatically outperforms probability
averaging. This led us to a hypothesis---inference speed is an implicit
confidence signal---and a method: winner-take-all training that internalizes
ensemble diversity within a single model.

Our results on Sudoku-Extreme are striking: a single 7M parameter model achieves
96.9\% $\pm$ 0.6\% puzzle accuracy (best seed: 97.6\%), matching a 3-model
halt-based ensemble and improving 11pp over the TRM baseline. The key insight is
not architectural but algorithmic: by letting $K{=}4$ parallel hypotheses
compete during training, we teach the model to find confident solutions faster.

The broader implication is that \textbf{how quickly} a model converges may be
as informative as \textbf{what} it outputs. This connects to a rich literature
on speed-accuracy tradeoffs in human cognition, where fast decisions correlate
with confidence. Adaptive computation mechanisms like ACT may be learning to
detect this same signal.

\ifdefined\isicml\vspace{-1em}\fi
\paragraph{Future directions.}
\begin{itemize}
  \item \textbf{Beyond Sudoku.} Validating on ARC, maze-solving, and other
    reasoning benchmarks.
  \item \textbf{Scaling $K$.} Understanding why $K{=}4$ saturates and whether
    harder domains require more heads.
  \item \textbf{Head specialization.} Encouraging heads to learn diverse
    strategies rather than converging to similar solutions.
  \item \textbf{Theoretical foundations.} Formalizing the connection between
    convergence speed and solution quality.
\end{itemize}


\bibliographystyle{plainnat}
\bibliography{references}

\newpage
\appendix

\section{Approaches That Failed}
\label{app:failed}

We document approaches that failed to improve over baseline, as negative results
inform future research directions. All experiments below achieved \emph{lower}
accuracy than standard TRM training.

\subsection{Diversity Through Training Losses}

Several approaches attempted to create diverse solution paths during training,
motivated by the success of halt-first ensembling at inference.

\paragraph{Forced trajectory divergence.} Adding a loss penalizing high cosine
similarity between trajectories from different input permutations. Despite
successfully reducing similarity (0.83 final), accuracy dropped from 86.3\% to
48.2\%---the model learned to satisfy both permutations with \emph{worse}
solutions rather than better diverse ones.

\paragraph{Inter-H diversity loss.} Penalizing consecutive H-iteration
similarity to prevent ``redundant'' iterations. This broke iterative
refinement: each step began undoing the previous step's work. Healthy models
show $\cos(z_H^{(h)}, z_H^{(h+1)}) > 0.9$; low similarity is harmful.

\paragraph{Multi-head output diversity.} Training $K{=}3$ separate value heads
on a shared trunk with diversity loss \emph{subtracting} similarity. Heads
collapsed to identical outputs within 1000 steps despite the diversity pressure.
The shared trunk creates a convergence attractor that overwhelms any loss signal.

\subsection{Architectural Diversity Mechanisms}

\paragraph{Slot attention.} Replacing $z_H$ with $K{=}3$ competing slots, forcing
cells to distribute information via softmax competition. Slots collapsed to
identical representations---the shared reasoning blocks (MLPMixer) dominate any
structural diversity mechanism.

\paragraph{K-specific full-rank weights.} Giving each of $K{=}3$ hypotheses
completely separate reasoning parameters (3$\times$ model size). Despite no
weight sharing, all hypotheses converged to the same solution. Separate weights
$\neq$ diversity; training dynamics must force divergence.

\paragraph{Learned $z_H$ initialization.} A puzzle-encoder MLP mapping inputs
to per-puzzle initial states. Accuracy dropped 1.9\%---fixed initialization
outperforms learned initialization, which adds noise and overfitting risk.

\paragraph{Multi-init training.} $K{=}3$ learned $(z_H, z_L)$ initialization pairs
with first-to-halt racing. Initial advantage vanished by step 2500; shared
weights caused all initializations to converge to the same solution.

\subsection{Loss Weighting and Curriculum}

\paragraph{Per-H label smoothing.} Softer targets early (exploration), sharper
late (commitment). Consistently 10+ percentage points behind baseline---high
average smoothing hurt convergence without compensating benefits.

\paragraph{Confidence-weighted loss.} Upweighting high-entropy cells. Initially
neutral but diverged late (--3\% at convergence); extra weighting destabilized
training.

\paragraph{Stuck-puzzle loss weighting.} 3$\times$ weight for puzzles where
$q_\text{halt} < 0.5$. With 91\% of puzzles marked ``stuck'' early in training,
this overwhelmed the loss landscape.

\subsection{Fixed-Point and Contraction Losses}

\paragraph{Fixed-point loss.} Penalizing $\|z_H^{(h+1)} - z_H^{(h)}\|$ to
encourage convergence. Accuracy dropped to ${\sim}65\%$. The model needs
continued iteration even when states are similar.

\paragraph{Validity loss.} Auxiliary loss encouraging predictions to satisfy
Sudoku constraints (no duplicates per row/col/box). All variants (symmetric,
low-temperature, Gumbel) failed at ${\sim}65\%$ accuracy.

\subsection{Input Manipulation}

\paragraph{Input corruption.} Corrupting 50\% of given digits during training
(diffusion-inspired). Destroyed learning---the model needs constraint information
intact. This contrasts with corrupting \emph{predictions} ($z_H$ noise), which
helps marginally.

\paragraph{Stochastic cell selection.} Updating random 50\% of cells per H-step
(ConsFormer-inspired). Failed by --3\%: TRM needs all cells to propagate
constraints simultaneously.

\subsection{Inference-Time Mechanisms at Training}

\paragraph{Binary reset during training.} Resetting $z_H/z_L$ when
$q_\text{halt} < 0.5$. DOA at step 500---destructive gradient signals from 41\%
of samples stuck at maximum iterations.

\paragraph{Entropy regularization.} Encouraging uncertainty at early H-steps.
Accuracy dropped 0.5\%: the model benefits from committing early on easy cells
and propagating, not from hedging.

\subsection{Architecture Modifications}

\paragraph{Causal attention.} Replacing MLPMixer with causal TransformerBlock.
Failed at --28\% behind baseline: TRM requires \emph{bidirectional} constraint
propagation. Causal masking prevents cell 80 from seeing cell 0, breaking
Sudoku's global structure.

\paragraph{Larger model.} Increasing hidden dimension from 512 to 768. Unstable
training with eventual collapse---the MLPMixer architecture doesn't benefit from
extra capacity on this task.

\subsection{Summary}

\begin{table}[h]
\centering
\small
\begin{tabular}{lc}
\toprule
Approach & $\Delta$ vs Baseline \\
\midrule
Forced trajectory divergence & --38\% \\
Causal attention & --28\% \\
Inter-H diversity loss & --20\% \\
Multi-head diversity loss & --15\% \\
Slot attention & --12\% \\
Per-H label smoothing & --11\% \\
Stochastic cell selection & --3\% \\
Confidence-weighted loss & --3\% \\
K-specific full-rank & --2\% \\
Learned $z_H$ init & --2\% \\
Entropy regularization & --0.5\% \\
\bottomrule
\end{tabular}
\caption{Summary of failed approaches, sorted by accuracy drop.}
\label{tab:failed}
\end{table}

The common failure mode: approaches that work for other architectures
(diffusion, ViTs, LLMs) often fail for TRM because iterative refinement has
different dynamics than single-pass inference or autoregressive generation.
TRM needs:
\begin{enumerate}
  \item Bidirectional information flow (no causal masking)
  \item High iteration-to-iteration similarity (incremental refinement)
  \item Early commitment on easy cells (propagation, not hedging)
  \item Intact input constraints (no corruption)
\end{enumerate}

The success of WTA training, in contrast, works \emph{with} these dynamics:
it creates diversity in the \emph{initialization} (where it matters) while
allowing each hypothesis to follow normal iterative refinement.

\end{document}